\documentclass[10pt,twocolumn,letterpaper]{article}

\usepackage[pagenumbers]{cvpr} 

\usepackage{graphicx}
\usepackage{amsmath}
\usepackage{amssymb}
\usepackage{booktabs}

\usepackage{dblfloatfix}
\usepackage[accsupp]{axessibility}  

\usepackage{kotex}
\usepackage{xcolor}

\usepackage[pagebackref,breaklinks,colorlinks]{hyperref}

\usepackage[capitalize]{cleveref}
\crefname{section}{Sec.}{Secs.}
\Crefname{section}{Section}{Sections}
\Crefname{table}{Table}{Tables}
\crefname{table}{Tab.}{Tabs.}

\def\cvprPaperID{20} 
\def\confName{CVPR}
\def\confYear{2022}

\begin{document}


\title{GP22: A Car Styling Dataset for Automotive Designers}


\author{Gyunpyo Lee \hspace{1cm} Taesu Kim \hspace{1cm} Hyeon-Jeong Suk\\
Korea Advanced Institute of Science and Technology\\
Color Lab, Department of Industrial Design\\
{\tt\small [gyunpyolee, tskind77, color]@kaist.ac.kr}
}
\maketitle

\begin{abstract}
An automated design data archiving could reduce the time wasted by designers from working creatively and effectively. Though many datasets on classifying, detecting, and instance segmenting on car exterior exist, these large datasets are not relevant for design practices as the primary purpose lies in autonomous driving or vehicle verification. Therefore, we release GP22, composed of car styling features defined by automotive designers. The dataset contains 1480 car side profile images from 37 brands and ten car segments. It also contains annotations of design features that follow the taxonomy of the car exterior design features defined in the eye of the automotive designer. We trained the baseline model using YOLO v5 as the design feature detection model with the dataset. The presented model resulted in an mAP score of 0.995 and a recall of 0.984. Furthermore, exploration of the model performance on sketches and rendering images of the car side profile implies the scalability of the dataset for design purposes. 


\end{abstract}

\vspace{-5mm}

\section{Introduction}
\label{sec:intro}

Marriage between refined styling and engineering is an essential aspect of vehicle design amongst all the other product designs. To aid such high proficiency required upon designers, designers are skilled in visualization and realization of the design in both manual and digital and knowledged in understanding human ergonomics and vehicle package engineering \cite{gkikas2012automotive}. Furthermore, vehicle design or styling inherits the vehicle's type and purpose defined by vehicle packaging and safety regulation constraints—such aspects limiting the styling parameters for designers \cite{macey2009h}.

\begin{figure}[b]
\vspace{-2mm}
  \centering
  \begin{subfigure}{0.15\textwidth}
    \includegraphics[width=\textwidth]{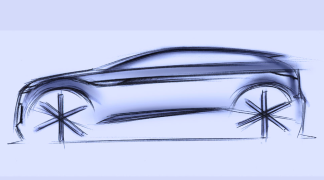}
    \caption{Ideation sketch}
    \label{fig:hair-stga}
  \end{subfigure}
  \hfill
  \begin{subfigure}{0.15\textwidth}
    \includegraphics[width=\textwidth]{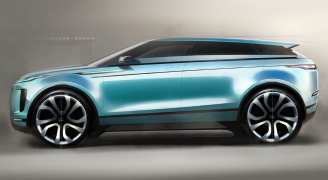}
    \caption{Sketch rendering}
    \label{fig:hair-stgb}
  \end{subfigure}
    \hfill
  \begin{subfigure}{0.15\textwidth}
    \includegraphics[width=\textwidth]{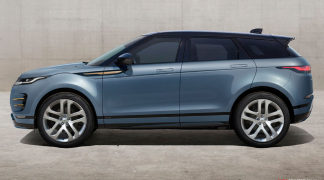}
    \caption{Actual vehicle}
    \label{fig:hair-stgc}
  \end{subfigure}
  \vspace{-2mm}
  \caption{Design process from car side profiles}
  \label{fig:hairystg}
  
\end{figure}

Within these inherited limitations, designers provide a tailored solution through styling for consumers with various preferences; the styling reinforces visual differentiation in perceiving the vehicle's form language. These characteristics are best found in the vehicle's side profile/vehicle side view \cite{luccarelli2014automotive}. It contains all the necessary determinants of design: the vehicle silhouette, proportion, and the entire image of the vehicle while bridging the brand identity characterized in the front and back of the vehicle (Figure \ref{fig:hairystg}). For this reason, the vehicle side view is one of the essential areas in vehicle design.

Many visual ideas are created during the vehicle design process, and the method of supporting the designer's day-to-day design process in archiving design outputs with valuable design ideas is overlooked \cite{ranscombe2011characterizing}. Hence, designers often reproduce similar proposals or search into the old project folders to revisit them. Unfortunately, such activity results in less productivity and creativity in the entire design process. Therefore, there is a need for support for designers with design-oriented archiving methods while matching processes to the current technology development. 


Computer vision methods like image classification and object detection could provide a fertile foundation for building a scalable yet fully automatable design database. Computer vision in the automotive field has already found various methods to identify cars in response to autonomous driving capability and vehicle verification and identification. There are large datasets supporting it. For example, current image detection datasets related to cars are as follows:
1) Car identification/classification in makes, engine types, years produced \cite{yang2015large, li2018machine}, 
2) Car verification on the road \cite{wang2017orientation, song2019apollocar3d},
3) Car pose estimation on the road \cite{liu2016deepLearn, sochor2016boxcars},
4) Car component detection and segmentation in its parts \cite{liu2016deep, pasupa2021evaluation}.

The primary purpose of these datasets is to support vehicles in identifying road situations while in autonomous driving mode, identifying car parts, or recognizing whole cars for surveillance reasons. However, utilizing such a dataset for design purposes is limited as the pre-defined annotations, and the dataset's structure does not match the designer's needs. For this reason, design-relative rules in dataset creation are required.





This study introduces design taxonomy based on vital aspects of the vehicle side profile as a guideline and annotated dataset, GP22: a dataset relevant to car designers' design practices \footnote{\url{https://doi.org/10.5281/zenodo.6366808}}. The dataset involves side profiles of cars produced or presented since the 1960s (Figure \ref{fig:hair-stgc}). It delivers a foundation and guideline for a quantitative approach to explaining car styling features. Then, we created a design feature detection model with our dataset. The model could serve as a tool in collecting and organizing the styling data. 


Combining the two will contribute immensely to building and reinforcing an inspirational design database. Thus, such usage could be further implemented in the automotive design studios to create their design database to maintain high-quality outputs. Through this method, designer intuition could be backed by quantifiable data while reducing unnecessary waste of time for designers in their creative process during design development.

\section{GP22}
The section covers the overview of the GP22 dataset in collecting the vehicle profile images under vehicle segment and makes. In addition, it delivers the taxonomy of vehicle design features relevant to design practices and the annotation guideline. 


\subsection{Image Collection}
A total of 1,480 car side profile photos are collected in the dataset. It includes cars produced or shown in the auto shows as concept vehicles since the 1960s. Collected images are classified by their brands and car segments, a criterion for estimating the car type and size—a total of 37 automakers and ten car segment categories defined in the dataset. Every image is labeled with all of these classifiers. 


\vspace{1.5mm}
\noindent \textbf{Segment} The vehicle segment in the dataset is reorganized and defined by reviewing multiple references: US EPA car class \cite{USEPA}, Euro NCAP car class \cite{EuroNCAP68}, H-Point \cite{macey2009h}, the book on vehicle packaging and ergonomics, and SAE technical paper on car design features \cite{luccarelli2014automotive}. As a segment defines the body style of the car design, it is an essential element in identifying relevant stylings and hardpoints to consider. Therefore, the dataset defines the segment as follows: A, B, C, D, E, F, J, S, P, and M. 

A to F segment includes cars from a city car, such as Smart, to the full-size saloon vehicle, such as Bentley Flying Spur. A, B, and C segments include hatchbacks, while C, D, E, and F include wagons and fastbacks of the sedan category. J segment is for all types of Sports Utility vehicles (SUV), from Compact Utility vehicles (CUV) to off-roaders. Furthermore, the S segment covers sports and exotic vehicles such as Ferrari 488 GTB and convertible coupes like BMW Z4. P segment includes various kinds of pickup trucks. Lastly, the M segment includes Multi-Purpose vehicles, from mini-vans to full-size vans.


\vspace{1.5mm}
\noindent \textbf{Brand} The brand included in the dataset contains 37 automakers. There are various automakers and groups; we tried to provide the deep learning model to identify minuscule differences in the side profile. However, large automakers such as Toyota, Volkswagen, Mercedes Benz, and Hyundai-Kia are the majority in the dataset. 



\subsection{Labels}
\subsubsection{Design features}
\begin{figure}[h]
\vspace{-2mm}
    \centering
    \includegraphics[width = .475\textwidth]{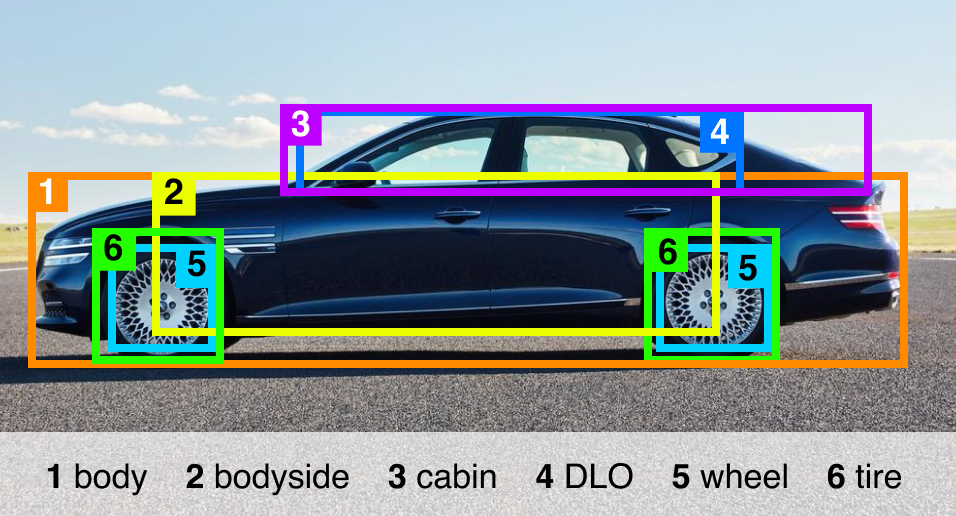}
    \caption{Figure of styling features and its labels}
    \label{fig:labelFig}
\vspace{-5mm}
\end{figure}

The essential styling features from the car side profile can be categorized with six key features: body, bodyside, cabin (greenhouse), daylight opening (DLO), wheel, and tire. It explains the car styling characteristics observed from the profile (Figure \ref{fig:labelFig}). 


\begin{figure*}[t]
    \centering
    \includegraphics[width=\textwidth]{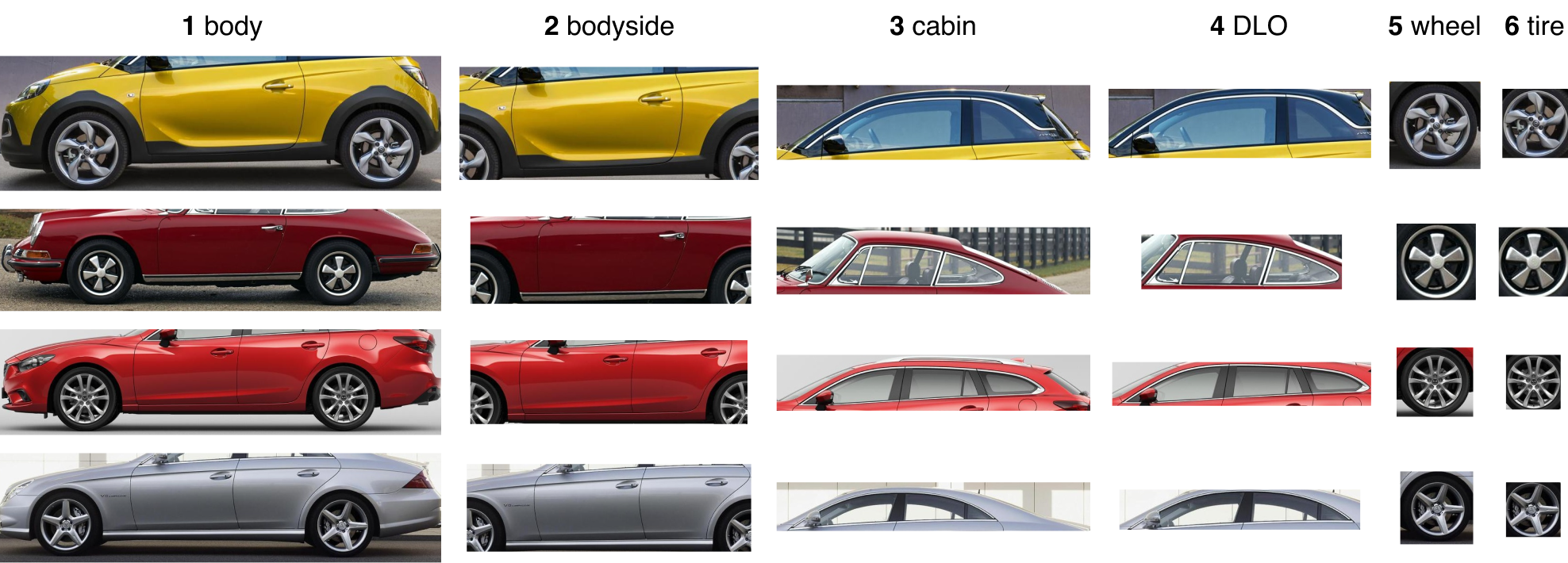}
    \caption{Use case example of the GP22 dataset as a design reference}
    \label{fig:overview}
    \vspace{-3mm}
\end{figure*}

\vspace{1.5mm}
\noindent \textbf{Body} The body defines the vehicle length from the front-end to the rear-end and height from the ground contact point to the daylight opening (DLO). It can be annotated by horizontally connecting the vehicle's front and back endpoint and vertically from the ground to the b-pillar area. 


\vspace{1.5mm}
\noindent \textbf{Bodyside} The bodyside is one of the vital areas in the car profile as it expresses car characteristics in its shapes and volumes. Connecting the area between a wheelbase, the length between the front wheel center and the rear wheel center,  and the wheelbase ground clearance to the b-pillar to annotate. It represents the dominant design area in the car profile.


\vspace{1.5mm}
\noindent \textbf{Cabin} The cabin communicates its passengers' habitability and visibility while expressing styling characteristics from the vehicle silhouette. This area starts from the cowl point, the frontmost point of the windshield, to the end of a rear wind glass where the trunk lid starts; it can be annotated correspondingly.


\vspace{1.5mm}
\noindent \textbf{Daylight opening (DLO)} The daylight opening (DLO) provides side openings for passengers. It also provides a unique design element from the car profile, like a famous BMW Hofmeister kink. Annotation can be done by connecting the front-end to the rear-end point of the DLO to define the length and the highest point of the DLO to define the height. 


\vspace{1.5mm}
\noindent \textbf{Wheel} The wheel provides a visual reference point for estimating the car's proportion, size, and characteristics. It also provides a perception of the stance of the car. Annotation can be done by wrapping each front and back wheelset from the profile.


\vspace{1.5mm}
\noindent \textbf{Tire} The tire is the primary reference point in checking the car wheelbase and other relative design elements of the car from the side. Annotation can be done by wrapping the entire tire area of each tire from front and back.

\vspace{1.5mm}

During the data annotation, upon mentioned design features and declared point of reference was taken into consideration for designers to annotate each side view images of the dataset formed by hand-picked car images. Therefore, participation of the designers in the data annotation process was essential as correctly locating the upon-mentioned design elements is the key to generating the designer's dataset. Furthermore, box annotation was chosen to provide ease in annotating and validating the labeled image (Figure \ref{fig:labelFig}).



\subsection{Design implication}
As an initial phase in building the design dataset, box extraction provides visual and analytical information necessary for designers in their design practice. Therefore, a properly labeled dataset following the above guideline can be utilized in two implications. First, it provides design feature visualization within relative styling characteristics. It can be done using the cropped images extracted from the original car side profile images (Figure \ref{fig:overview}). Furthermore, Detected boxes could be stored as a design database for archiving purposes either within the design studio or within the brand and could function as a mixed-initiative design data visualization for inspiration and reference purposes. As taxonomy elements represent design features of cars, they can be employed as reference materials in design effectively.

Another implication is utilizing box coordinates for the vehicle's proportion research by using tire length as a base metric for relatively understanding the design features (Figure \ref{fig:labelFig}). Key metrics found from the car side profile are wheelbase, front and rear overhang, dash-to-axle, and bodyside to DLO relationship. A front and rear overhang could provide the type of the vehicle, and it is achieved using body and bodyside relationship. Dash-to-axle implies the vehicle's status and drivetrain position, and it can be identified with DLO/cabin to the front endpoint of the body side. The longer the dash-to-axle, the car is perceived luxurious and prestigious. Lastly, the wheelbase and body to DLO relationship can provide information about the car's habitability and usability and vehicle package information. It can be observed using coordinates of bodyside and DLO heights.

\begin{figure*}[b]
\vspace{-3mm}
  \centering
  \begin{subfigure}{0.32\textwidth}
    \includegraphics[]{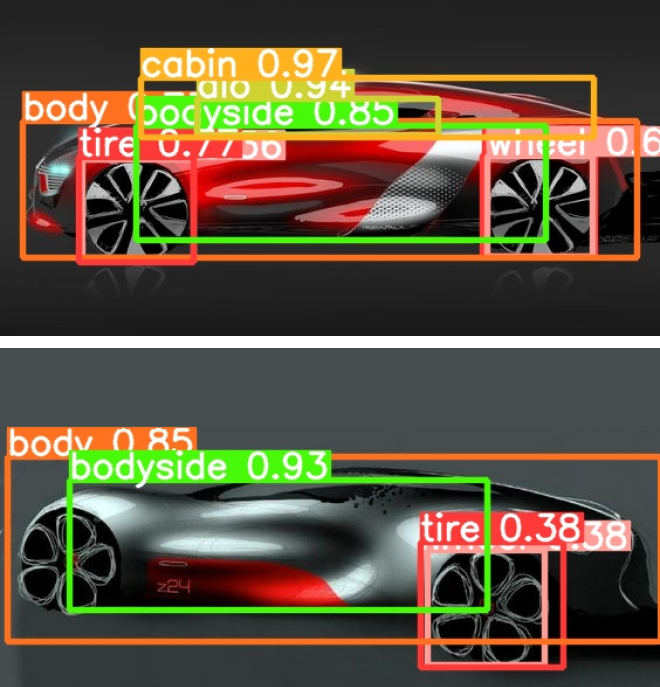}
    \caption{Sketch renderings}
    \label{fig:hair-a}
  \end{subfigure}
  \hfill
  \begin{subfigure}{0.32\textwidth}
    \includegraphics[]{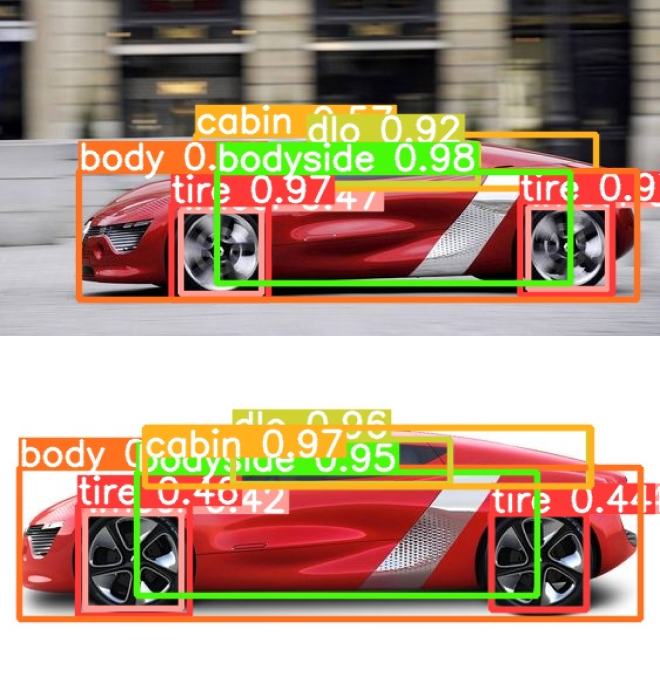}
    \caption{Actual cars}
    \label{fig:hair-b}
  \end{subfigure}
    \hfill
  \begin{subfigure}{0.32\textwidth}
    \includegraphics[]{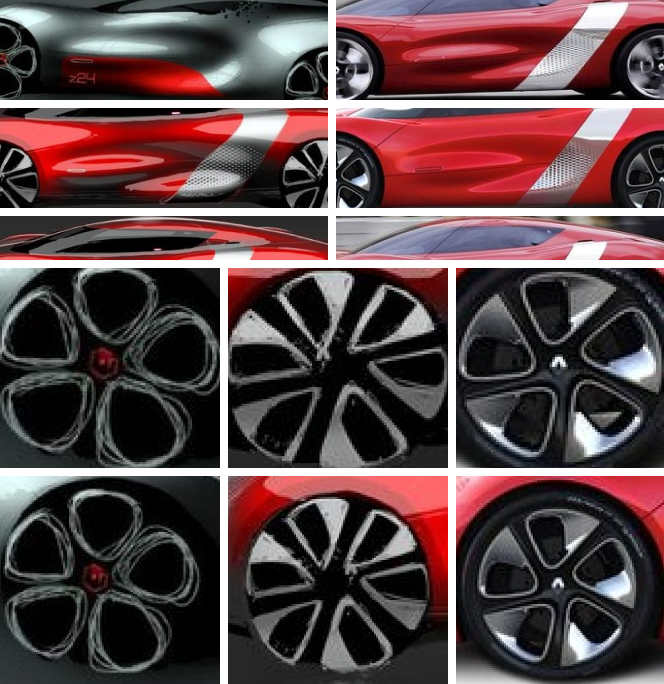}
    \caption{Cropped images}
    \label{fig:hair-c}
  \end{subfigure}
  \caption{YOLO v5 inference on official sideview sketch renderings, actual cars, cropped image for design development}
  \label{fig:hairy}
\end{figure*}

\section{Experiment}
This section introduces the process of generating a design feature detection model from the car side view by training and utilizing the YOLO v5 model \cite{glenn_jocher_2022_6222936} trained on the GP22 dataset. Then, we put car sketches and renderings as in-the-wild data on our trained model to analyze whether the trained model provides satisfactory results in extracting design features from unseen data. We also verify whether the model could support the design studio in constructing the design database. 



\subsection{Baseline model} There are many Convolutional Neural Network (CNN) models in object detection and image classification. Our model's objective is to create a design feature database from either design sketches or reference images in various sizes in the design studio environment. The study incorporates YOLO v5 \cite{glenn_jocher_2022_6222936} as a basis of the design feature detection model as it provides fast inference with reasonable accuracy in detecting objects while providing pre-trained models of diverse image sizes. Furthermore, the selected model is easily accessible for non-developers to train and use the model and such accessibility provides room for designers to get hands-on experience on deep learning algorithms compared to other renowned CNN models.




We performed transfer learning of the pre-trained model trained on different features. The GP22 dataset was split into 80:20 on each car segment category for the model's training. We used both car side profile photos and renderings not included in the dataset to test the model. Furthermore, we performed hyperparameter evolution to tune the hyperparameter in detecting design features more accurately. As a result shown in Table \ref{tab:results}, the mAP on validation of the base model was 0.989, the precision was 0.982, and the recall was 0.945. With the hyperparameter tuning, the mAP score increased from 0.989 to 0.995. Even though the precision decreased slightly from 0.982 to 0.953, the recall increased from 0.945 to 0.984: the high recall indicates that the model correctly identified and extracted design features from the given profile image, which fits our primary goal. 



\begin{table}[h]
\begin{tabular}{lccc}
\toprule
              &  mAP & Precision & Recall \\ \midrule
YOLOv5 (base)           &  0.989            & 0.982 & 0.945                 \\
YOLOv5 (hyperparameter) &  \textbf{0.995}  & 0.953 & 0.984                   \\ \bottomrule
\end{tabular}
\caption{Classification results}
\label{tab:results}
\vspace{-5mm}
\end{table}

\subsection{Model application on sketches and renderings}
To explore the model application, we used a test dataset that contains side profile sketches and photos of the same concept car, Renault Dezir. This test was to explore its performance on car sketches and renderings, as sketches are a means of communication between designers and designers-to-engineers in the studio environment. Thus, the result function as a determinant of the model is a stepping stone for developing design-relevant databases. 


The Figure \ref{fig:hairy}  shows that the model detected car design features from sketches correctly, even from the early ideation sketches of Renault Dezir that are more expressive in their style of rendering. Such performance can be applied to detect design elements and provide further inspiration in design development to transfer the impression created from the sketch onto the model. Furthermore, the result indicates that sketches and renderings of a similar view could also be used as part of the dataset to create a design database  (Figure \ref{fig:hair-c}).



\section{Conclusion}
The study proposes annotation guidelines using the taxonomy of car design features of the car side profile in the eye of the automotive designer. We constructed the GP22 dataset, a  tailored dataset for car design, particularly on car side profile images. With the dataset, we also created the design feature detection model using YOLO v5, which yields an 0.995 mAP score and 0.984 recall score. Furthermore, we examined the model application on the hairy data, which are data composed of sketches and renderings of the concept car to simulate the design studio implementation. The GP22 and its model will elevate car designers' creativity in their design process and possibly present a starting point of design-relevant database generation.



\section{Acknowledgement}
This research was supported by the 4th BK21 through the National Research Foundation of Korea (NRF) funded by the Ministry of Education (MOE) (NO.4120200913638).

{\small
\bibliographystyle{ieee_fullname}
\bibliography{egbib}

\begin{thebibliography}{10}\itemsep=-1pt

\bibitem{USEPA}
40 cfr § 600.315-08 - classes of comparable automobiles.
\newblock
  \url{https://www.govinfo.gov/app/details/CFR-2015-title40-vol30/CFR-2015-title40-vol30-sec600-315-08}.
\newblock (Accessed on 03/18/2022).

\bibitem{EuroNCAP68}
Euro ncap - the european new car assessment programme.
\newblock \url{https://www.euroncap.com/en}.
\newblock (Accessed on 03/18/2022).

\bibitem{gkikas2012automotive}
Nikolaos Gkikas.
\newblock {\em Automotive ergonomics: driver-vehicle interaction}.
\newblock CRC Press, 2012.

\bibitem{glenn_jocher_2022_6222936}
Glenn Jocher, Ayush Chaurasia, Alex Stoken, Jirka Borovec, NanoCode012, Yonghye
  Kwon, TaoXie, Jiacong Fang, imyhxy, Kalen Michael, Lorna, Abhiram V, Diego
  Montes, Jebastin Nadar, Laughing, tkianai, yxNONG, Piotr Skalski, Zhiqiang
  Wang, Adam Hogan, Cristi Fati, Lorenzo Mammana, AlexWang1900, Deep Patel,
  Ding Yiwei, Felix You, Jan Hajek, Laurentiu Diaconu, and Mai~Thanh Minh.
\newblock {ultralytics/yolov5: v6.1 - TensorRT, TensorFlow Edge TPU and
  OpenVINO Export and Inference}, Feb. 2022.

\bibitem{li2018machine}
Baojun Li, Ying Dong, Zhijie Wen, Mingzeng Liu, Lei Yang, and Mingliang Song.
\newblock A machine learning--based framework for analyzing car brand styling.
\newblock {\em Advances in Mechanical Engineering}, 10(7):1687814018784429,
  2018.

\bibitem{liu2016deep}
Hongye Liu, Yonghong Tian, Yaowei Yang, Lu Pang, and Tiejun Huang.
\newblock Deep relative distance learning: Tell the difference between similar
  vehicles.
\newblock In {\em Proceedings of the IEEE/CVF Conference on Computer Vision and
  Pattern Recognition}, pages 2167--2175, 2016.

\bibitem{liu2016deepLearn}
Xinchen Liu, Wu Liu, Tao Mei, and Huadong Ma.
\newblock A deep learning-based approach to progressive vehicle
  re-identification for urban surveillance.
\newblock In {\em Proceedings of the European Conference on Computer Vision},
  pages 869--884. Springer, 2016.

\bibitem{luccarelli2014automotive}
Martin Luccarelli, Markus Lienkamp, Dominik Matt, and Pasquale Russo~Spena.
\newblock Automotive design quantification: Parameters defining exterior
  proportions according to car segment.
\newblock In {\em SAE 2014 World Congress}, 2014.

\bibitem{macey2009h}
Stuart Macey.
\newblock {\em H-Point 2nd Edition: The Fundamentals of Car Design \&
  Packaging}.
\newblock 2009.

\bibitem{pasupa2021evaluation}
Kitsuchart Pasupa, Phongsathorn Kittiworapanya, Napasin Hongngern, and Kuntpong
  Woraratpanya.
\newblock Evaluation of deep learning algorithms for semantic segmentation of
  car parts.
\newblock {\em Complex \& Intelligent Systems}, pages 1--13, 2021.

\bibitem{ranscombe2011characterizing}
Charlie Ranscombe, Ben Hicks, Glen Mullineux, and Baljinder Singh.
\newblock Characterizing and evaluating aesthetic features in vehicle design.
\newblock In {\em Proceedings of the International Conference on Research into
  Design Engineering}, 2011.

\bibitem{sochor2016boxcars}
Jakub Sochor, Adam Herout, and Jiri Havel.
\newblock Boxcars: 3d boxes as cnn input for improved fine-grained vehicle
  recognition.
\newblock In {\em Proceedings of the IEEE/CVF Conference on Computer Vision and
  Pattern Recognition}, pages 3006--3015, 2016.

\bibitem{song2019apollocar3d}
Xibin Song, Peng Wang, Dingfu Zhou, Rui Zhu, Chenye Guan, Yuchao Dai, Hao Su,
  Hongdong Li, and Ruigang Yang.
\newblock Apollocar3d: A large 3d car instance understanding benchmark for
  autonomous driving.
\newblock In {\em Proceedings of the IEEE/CVF Conference on Computer Vision and
  Pattern Recognition}, pages 5452--5462, 2019.

\bibitem{wang2017orientation}
Zhongdao Wang, Luming Tang, Xihui Liu, Zhuliang Yao, Shuai Yi, Jing Shao,
  Junjie Yan, Shengjin Wang, Hongsheng Li, and Xiaogang Wang.
\newblock Orientation invariant feature embedding and spatial temporal
  regularization for vehicle re-identification.
\newblock In {\em Proceedings of the IEEE International Conference on Computer
  Vision}, pages 379--387, 2017.

\bibitem{yang2015large}
Linjie Yang, Ping Luo, Chen Change~Loy, and Xiaoou Tang.
\newblock A large-scale car dataset for fine-grained categorization and
  verification.
\newblock In {\em Proceedings of the IEEE/CVF Conference on Computer Vision and
  Pattern Recognition}, pages 3973--3981, 2015.

\end{thebibliography}
}

\end{document}